%% file: ReviewTemplate.tex
\documentclass[10pt,twocolumn,letterpaper]{article}

\usepackage[pagenumbers]{cvpr} 

\usepackage[table,xcdraw]{xcolor}
\usepackage{graphicx}
\usepackage{amsmath}
\usepackage{amssymb}
\usepackage{booktabs}
\usepackage{longtable}
\usepackage[pagebackref,breaklinks,colorlinks]{hyperref}
\usepackage[capitalize]{cleveref}
\crefname{section}{Sec.}{Secs.}
\Crefname{section}{Section}{Sections}
\Crefname{table}{Table}{Tables}
\crefname{table}{Tab.}{Tabs.}

\usepackage{multirow}

\def\cvprPaperID{232}
\def\confName{CVPR}
\def\confYear{2022}

\begin{document}

\title{Homophone Reveals the Truth: A Reality Check for Speech2Vec}

\author{Guangyu Chen\\
Renmin University of China \\
{\tt\small hcs@ruc.edu.cn}
}
\maketitle

\begin{abstract}
Generating spoken word embeddings that possess semantic information is a fascinating topic.
Compared with text-based embeddings, 
they cover both phonetic and semantic characteristics, which can provide richer information 
and are potentially helpful for improving ASR and speech translation systems.
In this paper,
we review and examine the authenticity of a seminal work in this field: Speech2Vec.
First,
a homophone-based inspection method is proposed to check the speech embeddings released by the author of Speech2Vec.
There is no indication that these embeddings are generated by the Speech2Vec model.
Moreover, through further analysis of the vocabulary composition, we suspect that a text-based model fabricates these embeddings.
Finally, we reproduce the Speech2Vec model, referring to the official code and optimal settings in the original paper.
Experiments showed that this model failed to learn effective semantic embeddings.
In word similarity benchmarks, 
it gets a  correlation score of 0.08 in MEN and 0.15 in WS-353-SIM 
tests, which is over 0.5 lower than those described in the original paper.
Our data and code are available\footnote{\url{https://github.com/my-yy/s2v_rc}}.
\end{abstract}


\section{Introduction}


Word embeddings are fixed-length representations of words,
which carry syntactic and semantic information 
and become the building block in natural language processing tasks,
such as named entity recognition~\cite{ner} 
and question answering~\cite{question_answer}.
Speech, as another carrier of information, also reflects semantic meaning,
which implies the possibility of directly using it to learn semantic word representations.
To this end, some attempts have been made to generate semantic speech embeddings~\cite{speech2vec2018,A18_CAWE,A43_chen2018phonetic,s2v_early2017}.
Compared with learning text-based embeddings, this is a much more difficult task.
Because there are intrinsic interferences between phonetics and semantics~\cite{A43_chen2018phonetic}.
For example, the words `ate' and `eight' have almost the same pronunciation but are semantically different.
If the model can not distinguish these sound-alike inputs,
 it is impossible to obtain effective semantic representations.
 
\begin{figure}[t]
	\includegraphics[width=7cm]{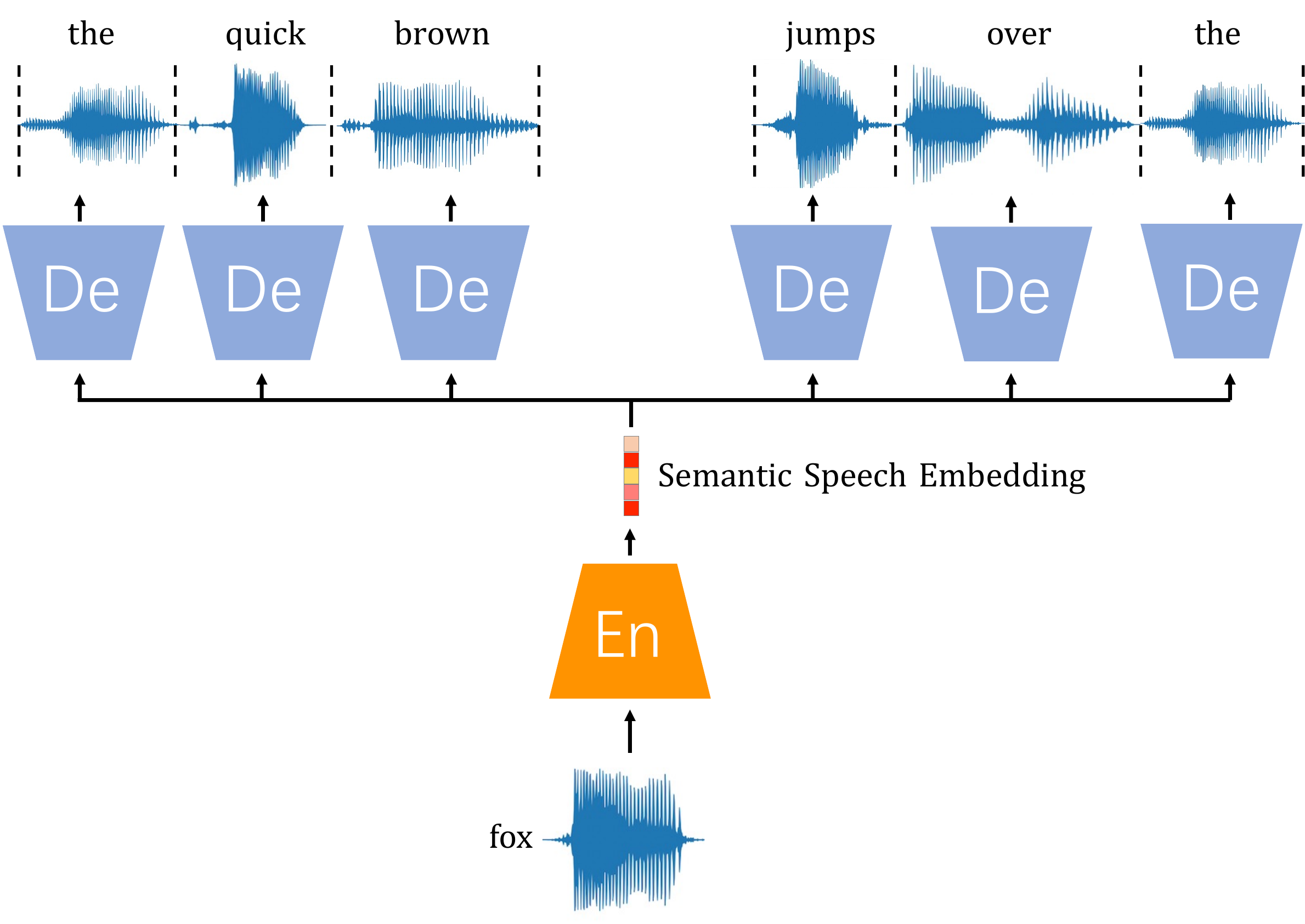}
	\centering
	\caption{ The Speech2Vec model (skip-gram style). 
	} 
	\label{fig:structure}
\end{figure}
In this paper,
we focus on Speech2Vec~\cite{s2v_early2017,speech2vec2018}, 
one of the field's earliest and most influential studies.
It can be viewed as a speech version of Word2Vec~\cite{word2vec},
which leverages the skip-gram and CBOW strategy to learn semantic speech embeddings.
Despite being an imitator of Word2Vec, 
it seems to have magically resolved the mentioned phonetic-semantic interference 
and has been reported to exceed the Word2Vec on 13 word similarity benchmarks~\cite{benchmark13}.
This phenomenon aroused our interest in exploring the validity of this study.
Specifically, we analyze the authenticity of Speech2Vec from two perspectives of 
\emph{the official-released speech embeddings} and \emph{our reproduction results}.
The conclusions can be summarized as follows:

1) The official speech embeddings fail to reflect speech characteristics, 
implying that they are not generated by the Speech2Vec model.

2) Analysis of the vocabulary composition reveals that these released embeddings may come from a text-based model.

3) Our reproduction results show that the Speech2Vec model 
failed to generate valid semantic speech embeddings. 
Even after 500 epochs of training, 
its performance is still negligible in most benchmark metrics.

\section{Recap Speech2Vec}
\label{sec:recap}

 Speech2Vec can be regarded as a speech version of Word2Vec,
which relies on the skip-gram or CBOW strategy to learn semantic speech embeddings.
Herein, we focused on the skip-gram style of Speech2Vec,
which is reported to have better performance. \cref{fig:structure} shows its structure.

When training, the input speech sentence is segmented into spoken word sequences in advance.
Then the encoder network converts the current spoken word into an embedding vector.
And this vector is used for decoding context words.
In the corpus, 
the audio segments corresponding to the same text word  appear multiple times 
(with different speakers, environment, emotions, or talking speed).
To obtain a deterministic word representation, 
the average of all embeddings corresponding to the same text word is taken as the final embedding.

The official code of Speech2Vec is not publicly available, 
but the pre-trained speech embeddings are released on GitHub
\footnote{\url{https://github.com/iamyuanchung/speech2vec-pretrained-vectors}}.
These embeddings are described from a skip-gram style Speech2Vec model trained
on 500 hours of LibriSpeech~\cite{librispeech} dataset:
\begin{quote}
	... In this repository, we release the speech embeddings of different dimensionalities learned by Speech2Vec using \textbf{skip-grams} as the training methodology. The model is trained on a corpus consisting of about \textbf{500 hours of speech} from LibriSpeech (the clean-360 + clean-100 subsets) ...
\end{quote}
We further provide a fork to this repository
\footnote{\url{https://github.com/my-yy/speech2vec-pretrained-vectors}}.

\section{Reality Check}
This section aims to check the authenticity of the released speech embeddings. 
We analyze them from two perspectives: phonetic characteristics and vocabulary composition.

\subsection{Homophone Inspection}  
Homophones refer to words that have the same pronunciation but reflect different meanings (e.g., `ate, eight',` dear, deer', and `whine, wine').
Since these spoken words are consistent in pronunciation, 
in a speech recognition system, 
the speech context  and language model~\cite{lm2000} are usually required to distinguish them.
For the skip-gram style Speech2Vec,
its input is  a single spoken word that does not contain context information.
Therefore, this model should not be able to distinguish these homophones effectively.
And the embedding similarity of the homophone pair should have a higher value.
We used 307 homophone pairs\footnote{
	Sourced from: \url{http://www.singularis.ltd.uk/bifroest/misc/homophones-list.html}
}
and calculated their similarities. 
The results are shown in~\cref{tab:homo_sim}.
\input{tables/1_homophone_sim}

As we can see,
for the official embeddings,
their homophone similarities are counterintuitively lower than the random results.
It suggested that the official model may not be affected by phonetic similarity and can
successfully distinguish these highly similar inputs. 
However, our reproduced model did not `show the magic'.
Although it had  a  0.77 similarity for random pairs (which implies the embedding space is ill-distributed), it still obtained a higher similarity of 0.95 for the homophone pairs.

\input{tables/2_homophone_pari}
\input{tables/3_knn}

\cref{tab:homo_pari_example}  further gives some pair examples.
In this table, we introduce a `Raw'  metric to measure the similarity of audio input.
This metric is the  cosine similarity of  extracted features from a pre-trained wav2vec~\cite{wav2vec2019} model.

It is clear to see that most homophone pairs have a Raw score close to 1.0, 
which means there is a high degree of similarity between these homophone words.
And the wav2vec model failed to distinguish them.
In addition, our reproduced model has similar behavior, which gets an average similarity of 0.95.
However, the official embeddings get relatively lower similarities.
Taking the `ate, eight' pair as an example,  
the similarities of Raw and ours are both 1.00, 
while the official gives a 0.22 similarity score.

The k-nearest neighbor results  (\cref{tab:knn}) further reveals the anomaly of the official embedding.
Although the official's nearest neighbors reflect similar semantics, 
there are considerable differences in pronunciation.
For example, the nearest neighbors of `hail'  are `thunder, blast, thunders',
while its homophone counterpart (`hale') is ranked as high as 15,253.
In contrast, the homophone rank in our reproduction is 9.

The above anomaly behavior of the official embeddings can be explained by a bold hypothesis: the official embeddings come from a text-based model.
As in a text-based model (like Word2Vec),
words of different spellings are assigned with different coding values.
In the learning process, 
the model does not rely on pronunciation to identify words; 
therefore, it will not be disturbed by them.
Moreover,  because homophone pairs have different meanings, their similarities may be smaller than the random pairs.
This explains why the official results have smaller similarities than the random in~\cref{tab:homo_sim}.

\subsection{Vocabulary Analysis}

\subsubsection{Checking the Released  Vocabulary}

\input{tables/4_librispeech}
The Seepch2Vec paper claimed that
the  training is based on  500 hours of LibriSpeech~\cite{librispeech}, which contains 1,252 speakers.
We refer to this dataset as  Libri-Clean. Its statistics are  shown in~\cref{tab:libir_stat}.
We counted distinct words in Libri-Clean's transcripts and compared them with the official.
The results are shown in~\cref{tab:vocab_compare}.
\input{tables/8_vocab_compare}

We found that 1.6k official words  are not in the Libri-Clean.
If we only consider frequent words  of occurrence count $\ge$ 5,  this number rises to 10k.
However, 
if we use the entire LibriSpeech corpus and  filter out infrequent words, 
 the vocabulary composition is exactly the same as the official.
This proves that the released embeddings are derived from all LibriSpeech corpus, not the claimed `500 hours' data.

In addition, 
we believe that these embeddings are generated by a text-based model, which directly uses text transcripts for training.
Because the LibriSpeech dataset needs to be pre-processed for training,
each speech sentence should be aligned with the transcript to locate word boundaries.
However, this alignment process is not ideal, which may cause data loss.

In our analysis, we used the alignment results based on the widely used Montreal Forced Aligner~\cite{montreal}~\footnote{\url{https://github.com/CorentinJ/librispeech-alignments}}.
It aligned the entire corpus of 292,367 speech sentences and encountered 127 failures (0.04\% fail rate).
After removing these unaligned speeches, 
the corpus changes, which causes additional 17 missing words in the vocabulary.
Therefore, if the authors of Speech2Vec do not realize an ideal alignment, 
their vocab size will not equal 37,622 exactly.
Otherwise, their embeddings come from a text-based model.

\input{tables/5_missing_words}

\subsubsection{Checking The Paper Data}
Although the vocabulary size is not mentioned in the paper of Speech2Vec,
we can still find some clues from the word similarity benchmark results.
In the original paper, these benchmarks are used to evaluate the semantic information carried by embeddings.
Each benchmark contains a series of predefined word pairs with the corresponding oracle correlation.
In an early version of Speech2vec~\cite{s2v_early2017}, 
the author reported the `not found' pairs in these benchmarks~(\cref{tab:missing_word}).

If we use all the 66,721 words in Libri-Clean to construct the vocabulary,
there are still more not found pairs in these tests (except for SimVerb-3500).
For example, the original paper reported 122 not found pairs in the MEN test,
but we found 231.
This illustrates that the authors of Speech2Vec used a larger corpus other than the claimed 500h Libri-Clean.

\section{Our Reproduction }
\label{sec:our_rep}

\subsection{Reproduction Details}
\subsubsection{Code Implementation}
Our reproduction is based on the incomplete code provided by the author of Speech2Vec.
This code covers the model implementation and training logic while lacking the data pre-processing.
We complement the missing parts and train the Speech2Vec, referring to the optimal setting in 
the original paper.
Specifically, the embedding dimension is set as 50.
The window size is taken as 3.
The encoder is a single-layer bidirectional LSTM.
And the decoder is a single-layer unidirectional LSTM.
During the decoding process, 
the attention mechanism~\cite{luong2015attention} is used to reference the encoder's outputs (\cref{fig:model_deail} in the appendix gives a detailed overview of its structure).

\subsubsection{Training Details}
In accordance with Speech2Vec's paper, the Libri-Clean dataset is used for training.
Each speech sentence is segmented into spoken words in advance.
Then we filter out the low-frequency words that appear less than four times and finally get a vocabulary of size 30,788.
For each spoken word,
 we convert it into  13-dim MFCC  frames.

During the mini-batch construction, 
we set batch size as 4096 and 
trained the model for 500 epochs with SGD optimizers of 0.001 learning rate.
This training process took 8.4 days on an AMD 3900XT + RTX3090 device.
The output models of every epoch are saved, and the 500th epoch's model is used for evaluation.

\begin{figure*}[t]
	\includegraphics[width=18cm]{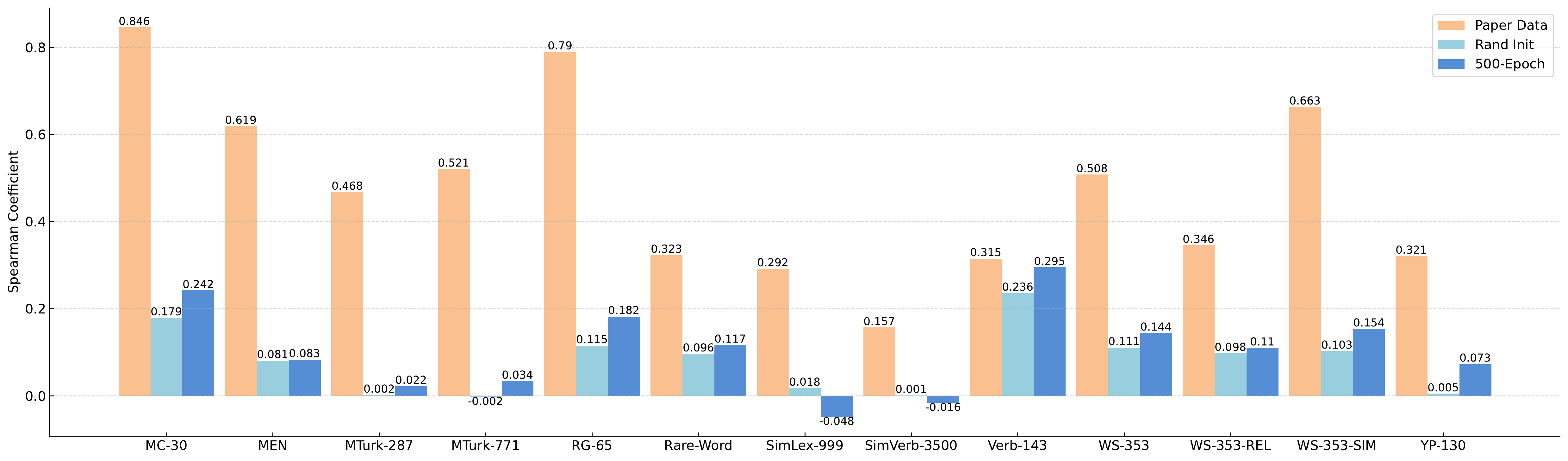}
	\centering
	\caption{
Comparisons on word similarity benchmarks. 
Paper Data: The results claimed in the paper of Speech2Vec~\cite{speech2vec2018}.
Rand Init: 
The randomly initialized model without training.
500-Epoch: Train the model for 500 epochs.	
} 
	\label{fig:score_compare}
\end{figure*}

\begin{figure*}
	\includegraphics[width=18cm]{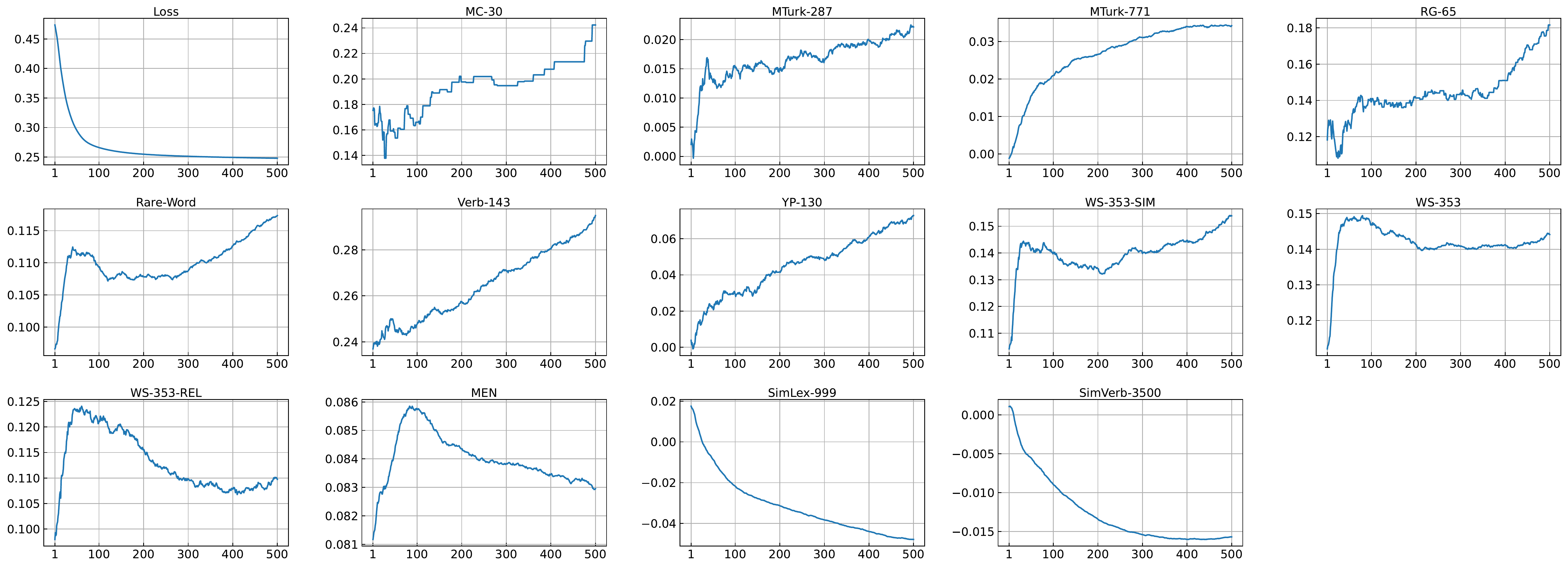}
	\centering
	\caption{ The learning curve.
		X-axis: training epochs. Y-axis:  Spearman coefficient (except for Loss).
	} 
	\label{fig:training_process}
\end{figure*}
\subsection{Results \&  Analysis}
Same with the official,
we tested our reproduced model on 13 similarity benchmarks~\cite{benchmark13}.
The evaluation score is Spearman's rank correlation coefficient $\rho$.
It has a range of [-1,1], and a larger value represents a better result.
The results are shown in~\cref{fig:score_compare}.

Obviously, the reproduced model gives trivial results on these benchmarks.
Compared with the random initialization, 
the 500-epoch model only has minor improvements 
or even decreases in indicators of SimLex-999 and SimVerb-3500.
Moreover, compared with the claimed paper data, 
there is a substantial performance drop, 
even over 0.5 in some metrics (MC-30, MEN, RG-65, WS-353-SIM).

We further depicted the training process in~\cref{fig:training_process}.
With the loss decreasing, most indicators showed a rising trend.
But the increasing speed is relatively slow and is finally limited to a small range (e.g., 0$\sim$0.022 for MTurk-287).
More training epochs may provide better results on these metrics, 
but it should be much larger than the claimed `500 epochs'.
Moreover, two indicators seem to have passed their peak period, with performance turning from rising to falling (WS-353-REL and MEN).
However, their peak performance is still relatively lower than the original paper.
There are two indicators whose performance is consistently declining,
which means that the required semantics have not been learned successfully (SimLex-999 and SimVerb-3500).

\begin{figure}
	\centering
	\begin{subfigure}{0.235\textwidth}
		\includegraphics[width=\textwidth]{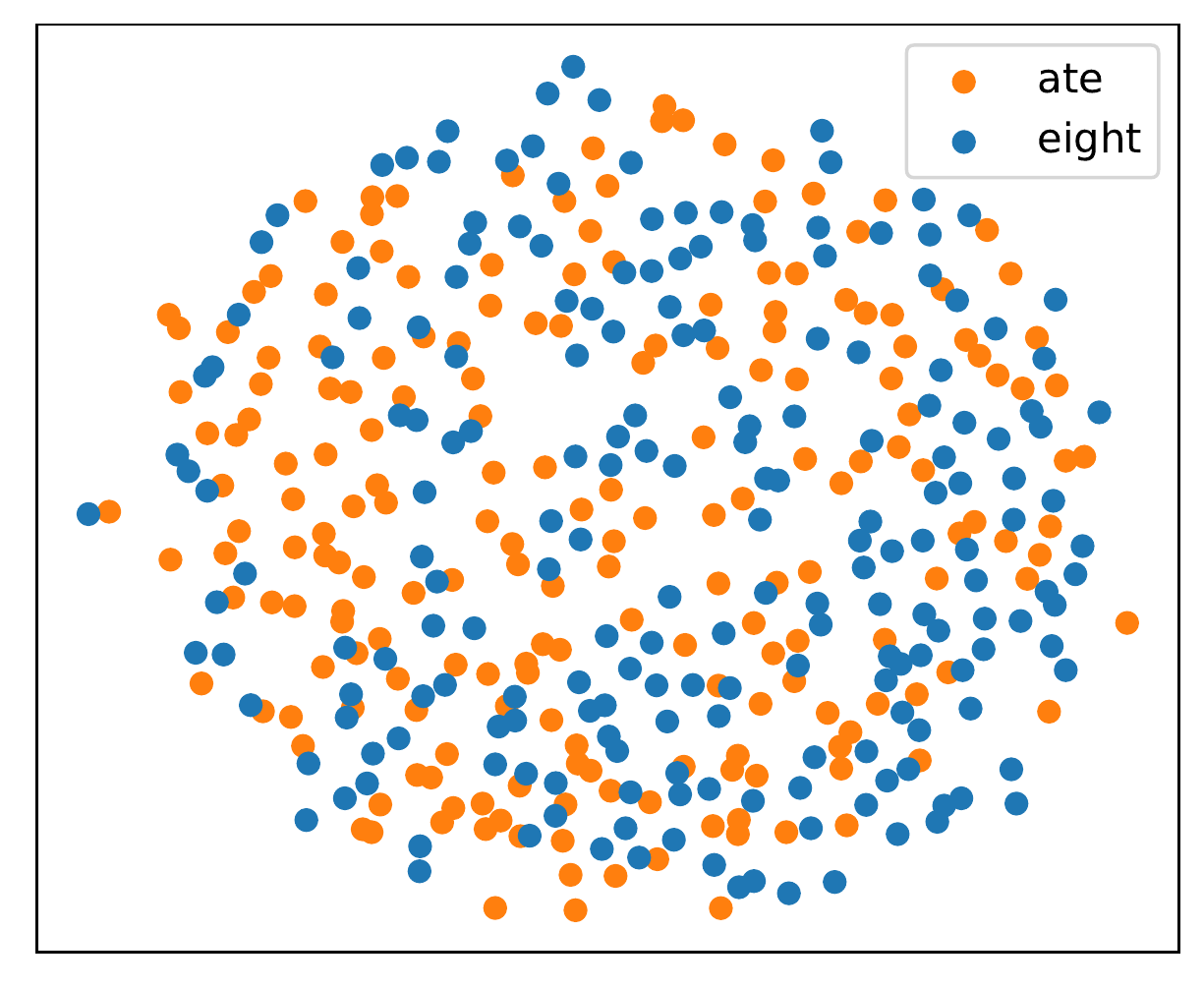}
	\end{subfigure}
	\hfill
	\begin{subfigure}{0.235\textwidth}
		\includegraphics[width=\textwidth]{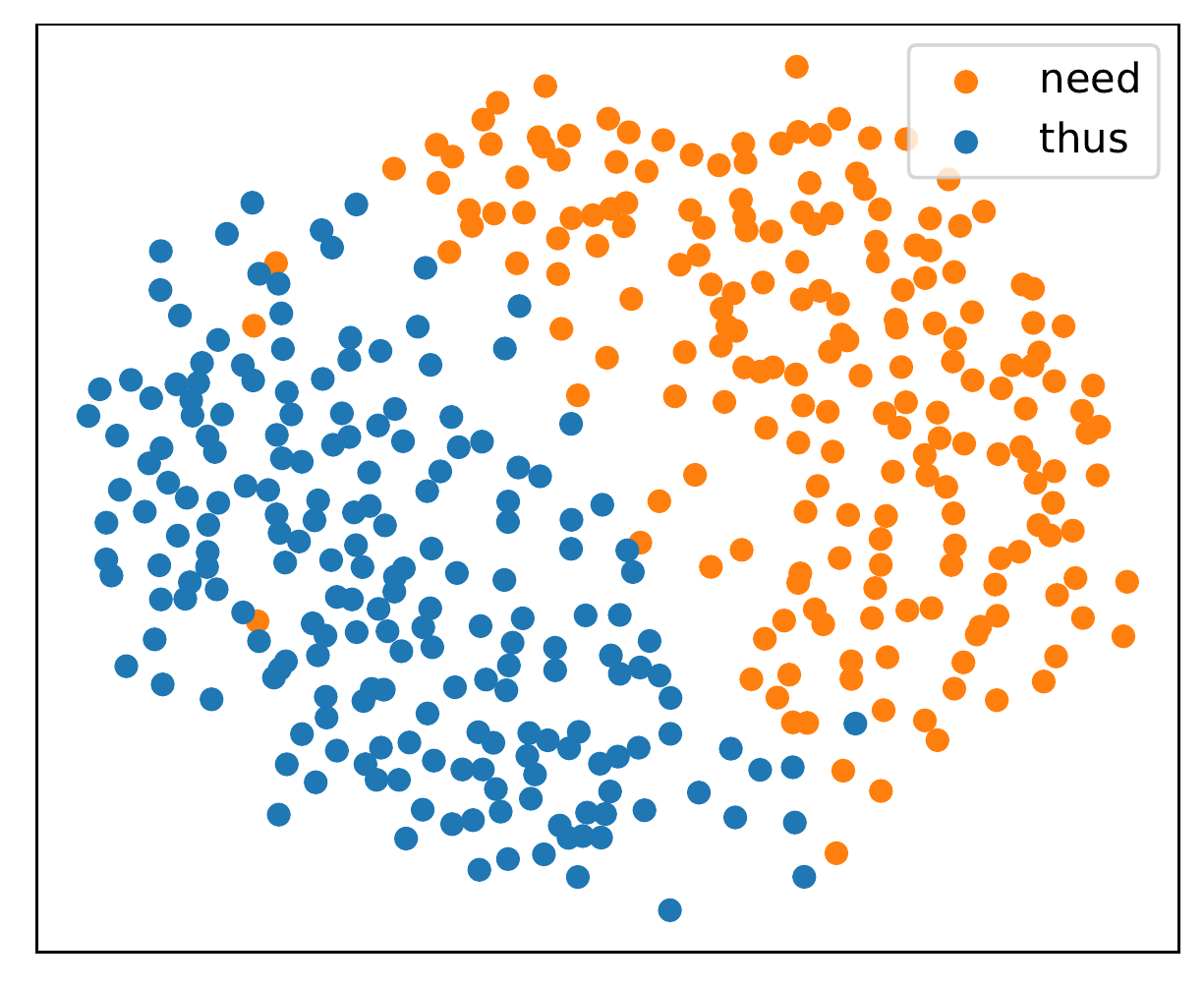}
	\end{subfigure}
	\hfill
	\begin{subfigure}{0.235\textwidth}
		\includegraphics[width=\textwidth]{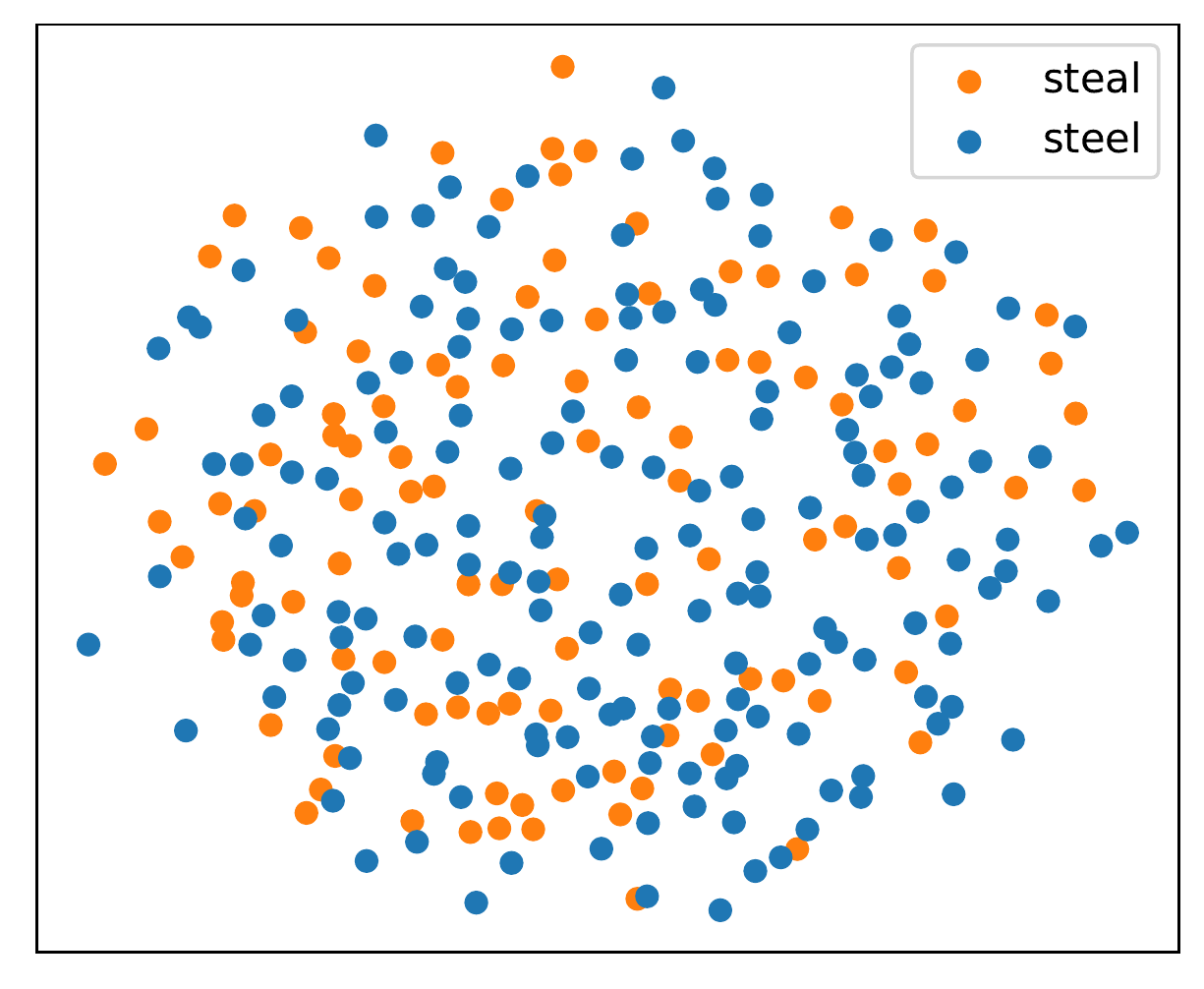}
	\end{subfigure}
	\hfill
	\begin{subfigure}{0.235\textwidth}
		\includegraphics[width=\textwidth]{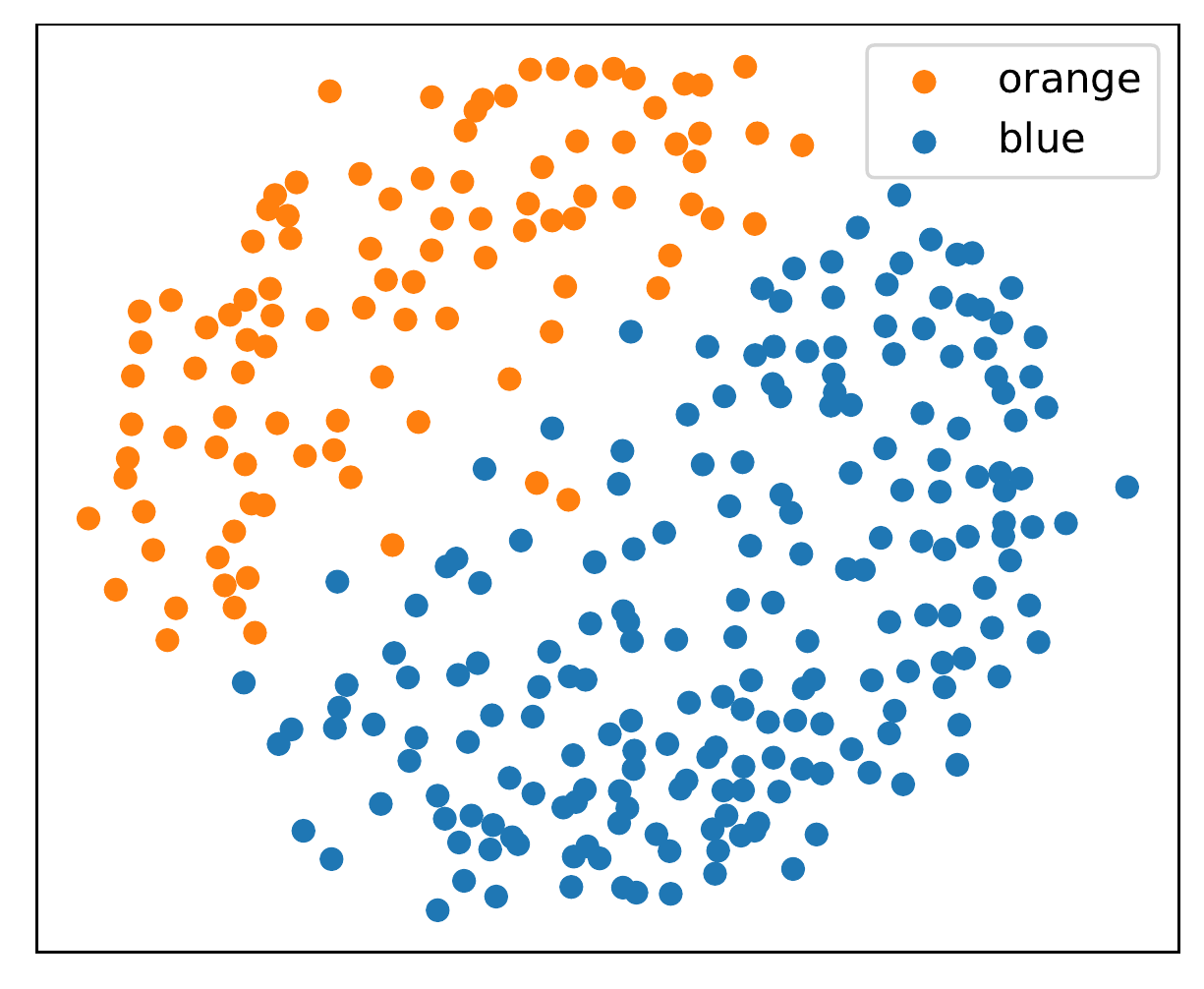}
	\end{subfigure}
	\caption{MDS visualization of speech word embeddings.}
	\label{fig:mds}
\end{figure}

As mentioned in~\cref{sec:recap}, each text word corresponds to multiple speech segments.
In~\cref{fig:mds}, we visualize the embedding distribution of specific words.
Specifically, the multi-dimensional scaling (MDS)~\cite{MDS2003} 
is used for visualization.
Compared with the popular T-SNE, the MDS preserves distance and global structure.

As we can see, the homophone words are mixed, 
while words with different pronunciations tend to be separated.
This result confirms that Speech2Vec's encoder can not successfully distinguish homophones.
Moreover, we noticed some mixed points for the `need-thus' pair.
This shows that the model did not have a good discernibility even for not sound-alike pairs,  
which may be caused by the interference between phonetics and semantics.

\section{Related Works}
\subsection{Works on Learning Semantic Speech Embeddings}
Five years since the Speech2Vec was proposed, 
there is rarely work in the field of learning semantic speech embeddings.
The Speech2Vec has received 158 citations so far, 
but no article has directly compared with it,
and no subsequent work has proved its validity.

\subsubsection{Chen et al.'s Work}
Chen et al.~\cite{A43_chen2018phonetic} pointed out that
when learning semantic speech embeddings, the phonetics and semantics inevitably disturb each other.
Therefore, they designed a two-stage learning scheme to reduce the learning difficulty.
In the first stage, 
an encoder-decoder structure generates phonetic embeddings in which the speaker characteristics are disentangled.
In the second stage, these embeddings possess semantic meaning through a context prediction task.
Experiments have verified their embeddings' phonetic and semantic characteristics. 
However, this work is not compared with Speech2Vec on the similarity benchmarks.

\subsubsection{CAWE}
CAWE~\cite{A18_CAWE} is an  encoder-decoder-based speech recognition model,
which relies on an attention mechanism to automatically segment speech sentences into spoken words and generate speech embeddings. 
This model is tested on 16 sentence evaluation tasks and shows competitive results to Word2Vec.
Compared with skip-gram style Speech2Vec,
the CAWE can leverage the context information.
Thus their effectiveness can not laterally prove Speech2Vec's validity.
Moreover, we have not found any successful third-party replications of CAWE.

\subsubsection{Audio2Vec}
Audio2Vec~\cite{A76_Audio2Vec} is a CNN-based model which uses skip-gram, CBOW, and temporal-gap strategies to generate fixed-length audio representations.
Although it has a similar training style to Speech2Vec, 
the goal of Audio2Vec is to produce general-purpose audio representations.
In the training process, the speech sentence is segmented into fixed-length slices without reference to word boundaries.
Thus this model was not tested on the word similarity benchmarks and not compared with Speech2Vec.

\subsection{Works  Directly Based on Speech2Vec Embedding}
Ideally, if the experimental data is questionable, the conclusion's validity cannot be guaranteed.
We note that there are some works~\cite{A27_alignment_speech_text,A7_chung2019towards,A8_punctuation_prediction} directly based on the embeddings of Speech2Vec.
Among them, two works come from the authors of Speech2vec ~\cite{A27_alignment_speech_text,A7_chung2019towards},
which align the speech space and text space to realize cross-language speech-to-text translation.

Yi et al.~\cite{A8_punctuation_prediction} reported that 
they successfully improved punctuation prediction performance by 
involving the released Speech2Vec embeddings in their input.
But they did not explore how these speech embeddings work.
We think their results are authentic but unintentionally provide a wrong interpretation.
As we have analyzed, these Speech2Vec embeddings cannot reflect phonetic information.
Thus their improvements come from involving more input features.
And it has nothing to do with whether these features come from a text-based or speech-based model.
Therefore, 
we believe that the above works cannot provide evidence for the authenticity of Speech2Vec.

\subsection{Replication Attempts by Other Researchers}
We investigated the replication attempts by other researchers, including
Jang's\footnote{\url{https://github.com/yjang43/Speech2Vec}}
and
Wang's\footnote{\url{https://github.com/ZhanpengWang96/pytorch-speech2vec}}.
Like us, they fail to achieve the results claimed in Speech2Vec's paper.
Jang reported that the model  gets trivial results.
As for Wang's,
the reproduced model  is also vulnerable, which only  gets $\rho=0.08$  in WS-353-REL and $\rho=-0.15$ in YP-130.

\section{Disscussion}
``If I have seen further, it is by standing on the shoulders  of Giants. ——Isaac Newton,1675.''

Scientific progress builds on the innovations of predecessors.
If an influential innovation is falsified, 
it is not `the shoulder of a giant'.
Instead, it becomes a \emph{stumbling block} and brings huge obstacles to developing the field.
We think science is not a kind of religion, 
especially when it comes to computer science.
The authenticity of  innovation should depend on reproducibility 
rather than the reputation of the author, institution, or paper grade.

In 2018, a master's student was assigned to begin his academic research in an emerging field.
Unluckily, 
he took faith in a bogus study and get off the wrong road.
He was passionate about it, striving for years to achieve an impossible goal, 
and ended up with nothing.
Four years after setting out, he reflected on the failures of his life and wrote this article.
To prevent others from repeating his past, this is the value of this study.

{\small
\bibliographystyle{ieee_fullname}
\bibliography{egbib}
}

\clearpage 
\onecolumn 
\appendix
\section{Encoding-Decoding Details}
\begin{figure}
	\centering
	\begin{subfigure}{0.35\textwidth}
		\includegraphics[width=\textwidth]{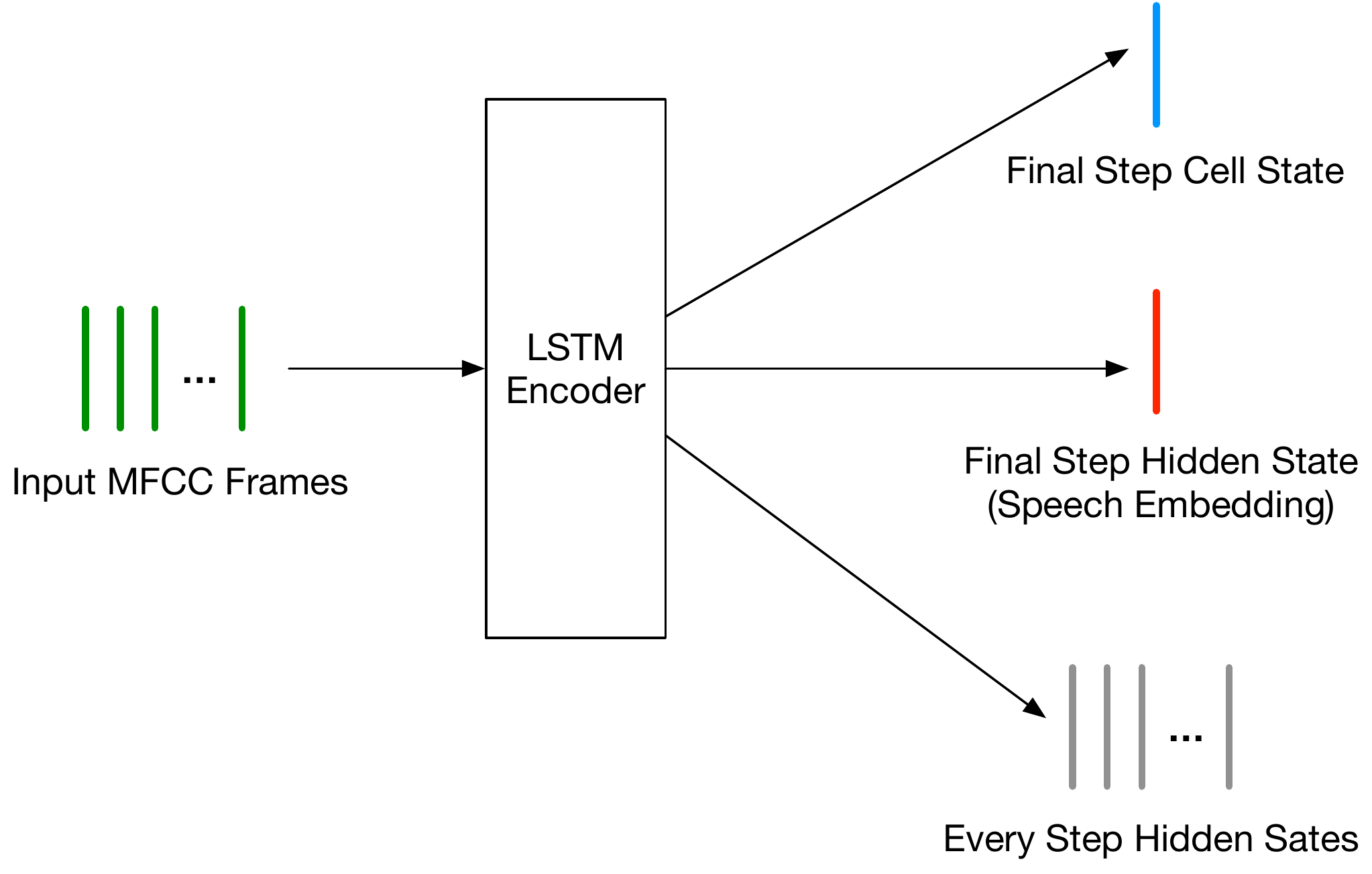}
		 \caption{Encoding process}
	\end{subfigure}
	\hfill
	\begin{subfigure}{0.62\textwidth}
		\includegraphics[width=\textwidth]{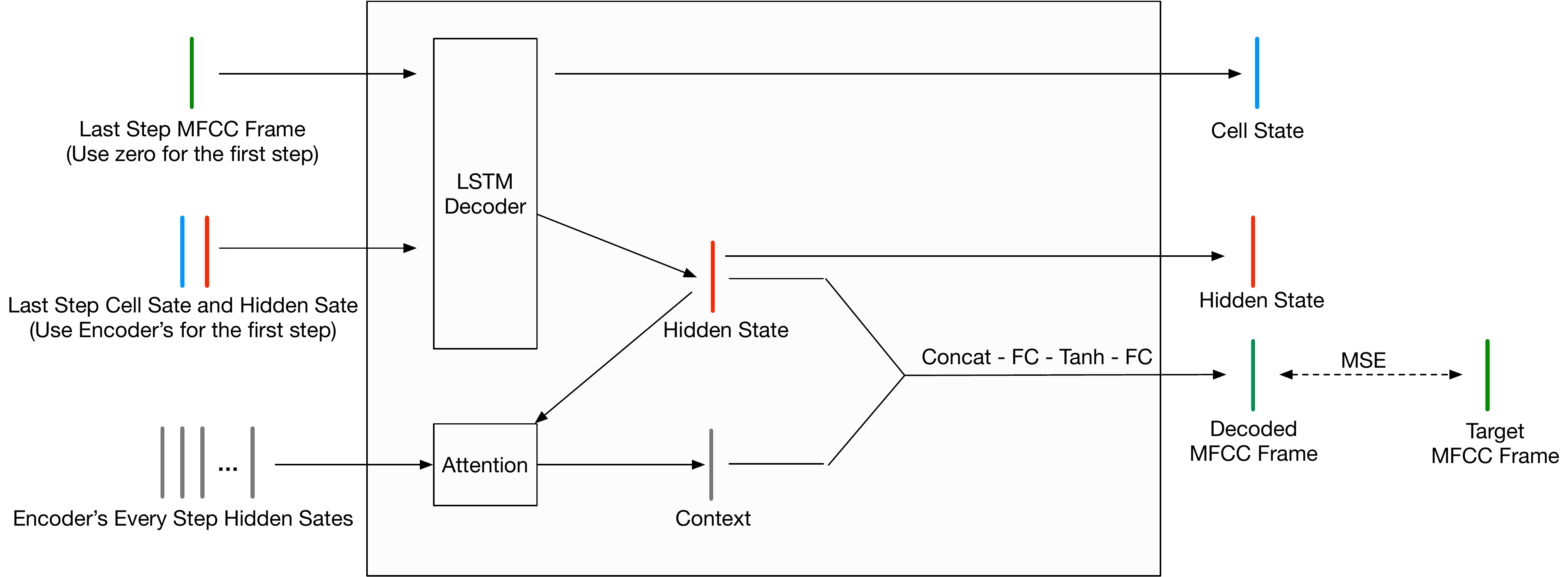}
		 \caption{One decoding step}
	\end{subfigure}
	\caption{
Illustration of the encoding-decoding steps. Best view in color. 
}
	\label{fig:model_deail}
\end{figure}

\cref{fig:model_deail} depicts the encoding-decoding process. 
For clarity, this illustration takes a single spoken word as input,
while the model is trained in batch mode.
Since the input words have various time lengths, 
the length uniformity is implemented in the official code.
The official code lacks the data processing part, and we don't know the exact time length.
According to statistics, our dataset has an average MFCC frame of 10, and there are 94.6\% of words less than 21 frames.
Thus we padded (or truncated) the input word to fixed 20 MFCC frames.

In the encoding stage, the official code used the \verb|pack_padded_sequence| of PyTorch to eliminate the padding effects.
However, in the decoding stage, 
the reconstruction loss corresponding to the padded frames is still considered.
We fixed this issue in our reproduction.
More details can be found in our released code.

\section{Comparisons on Similarity Benchmarks}
We note that the optimizer in the official code is the Adam optimizer 
(not the described SGD in the official paper).
To exclude the optimizer's impact, we tried the Adam setting.
However, we did not observe essential differences in the results, as shown in~\cref{tab:score_compare}.
\input{tables/6_score_compare}

\section{The Homophone List}
The full 307 homophone pairs are listed in~\cref{tab:homo_full_list}.
Since our reproduction is based on Libri-Clean,
which has a smaller vocab than the released embedding.
Thus there are 33 missing pairs in our reproduction (marked by `-').
\input{tables/9_full_homolist}

\end{document}

%% file: tables/1_homophone_sim.tex
\begin{table}[h]

	\centering

\begin{tabular}{c|c|c|c} 
	\toprule
	\multirow{2}{*}{Embedding}   & \multirow{2}{*}{Dim} & \multicolumn{2}{c}{Pair Similarity}  \\ 
	\cline{3-4}
	&                          & Homophone  & Random                  \\ 
	\hline
	\multirow{4}{*}{Official} & 50      & 0.33{\tiny ±0.15} & 0.39{\tiny ±0.17}\\
	& 100     & 0.28{\tiny ±0.14} & 0.35{\tiny ±0.15}\\
	& 200     & 0.23{\tiny ±0.12} & 0.31{\tiny ±0.14}\\
	& 300     & 0.21{\tiny ±0.11} & 0.29{\tiny ±0.14}\\
	\hline
Ours & 50 & 0.95{\tiny ±0.05} & 0.77{\tiny ±0.10}\\
	\bottomrule
\end{tabular}

	\caption{
		Cosine similarities of word pairs (mean ± std).
Random: Similarities between  307 randomly selected  pairs.
Official: Embeddings released by the author of Speech2Vec.
Ours: Embeddings generated by our reproduced model (detailly described in ~\cref{sec:our_rep}.
)
	} 
	\label{tab:homo_sim}
\end{table}

%% file: tables/2_homophone_pari.tex
\begin{table}[h]
	\centering
\begin{tabular}{ll|c|c|l} 
	\toprule
	\multicolumn{2}{l|}{\multirow{2}{*}{Homophone Pair}} & \multicolumn{3}{c}{Similarity}          \\ 
	\cline{3-5}
	\multicolumn{2}{l|}{}                                & Raw  & Official & \multicolumn{1}{c}{Ours}  \\ 
	\hline
ate & eight & 1.00 & 0.22 & 1.00 \\
hail & hale & 0.99 & 0.34 & 0.97 \\
know & no & 1.00 & 0.63 & 0.97 \\
made & maid & 1.00 & 0.25 & 0.98 \\
meat & meet & 1.00 & 0.13 & 0.99 \\
peace & piece & 1.00 & 0.15 & 0.99 \\
sea & see & 1.00 & 0.31 & 0.98 \\
steal & steel & 1.00 & 0.19 & 0.99 \\
tale & tail & 1.00 & 0.23 & 1.00 \\
whine & wine & 1.00 & 0.13 & 0.97 \\
	\hline
	\multicolumn{2}{c|}{MEAN}                            & 0.99 & 0.32     & 0.95                       \\
	\bottomrule
\end{tabular}
	\caption{
Examples of homophone  similarities. 
The Official stands for the official 50-dimensional embeddings.
The appendix gives the full 307 results.
	}
	\label{tab:homo_pari_example}
\end{table}

%% file: tables/3_knn.tex
\begin{table*}[t]
	\centering
\begin{tabular}{l|lll|c|lll|c} 
	\toprule
	\multirow{2}{*}{Word} & \multicolumn{4}{c|}{Official}                    & \multicolumn{4}{c}{Ours}                      \\ 
	\cline{2-9}
	& \multicolumn{3}{c|}{K Nearest Neighbor} & HR     & \multicolumn{3}{c|}{K Nearest Neighbor} & HR  \\ 
	\hline
ate	&	eating	&	drank	&	eat	&	25,099	&	eight	&	agent	&	aid	&	1 \\
hail	&	thunder	&	blast	&	thunders	&	15,253	&	hell	&	handle	&	held	&	9 \\
know	&	tell	&	think	&	why	&	2,305	&	narrow	&	natural	&	no	&	3 \\
made	&	making	&	gave	&	make	&	35,582	&	me	&	mid	&	necessity	&	15 \\
meat	&	roasting	&	roasted	&	venison	&	33,108	&	meet	&	mate	&	mute	&	1 \\
peace	&	tranquility	&	liberty	&	deliverer	&	33,609	&	plates	&	place	&	patience	&	7 \\
sea	&	ocean	&	shore	&	waters	&	18,610	&	city	&	safety	&	succeed	&	100 \\
steal	&	flay	&	stealing	&	kill	&	29,506	&	still	&	steel	&	special	&	2 \\
tale	&	story	&	adventure	&	tales	&	28,617	&	tail	&	title	&	tackle	&	1 \\
whine	&	snarling	&	growls	&	growl	&	34,575	&	watering	&	washing	&	wide	&	4 \\
	\hline
	MEAN                  &             &          &                & 25,626  &           &          &                  & 14  \\
	\bottomrule
\end{tabular}
	\caption{The k nearest neighbor results (k=3). HR: Homophone ranking.} 
	\label{tab:knn}
\end{table*}

%% file: tables/4_librispeech.tex
\begin{table}[h]

	\centering

\begin{tabular}{c|l|r|c|r|c} 
	\toprule
	\multicolumn{2}{c|}{Corpus Division}              & \multicolumn{2}{c|}{Hours}   & \multicolumn{2}{c}{Speakers}    \\ 
	\hline
	\multirow{4}{*}{Clean} & train-clean-100 & 100.6 & \multirow{4}{*}{475} & 251   & \multirow{4}{*}{1,252}  \\
	& train-clean-360 & 363.6 &                      & 921   &                         \\
	& dev-clean       & 5.4   &                      & 40    &                         \\
	& test-clean      & 5.4   &                      & 40    &                         \\ 
	\hline
	\multirow{3}{*}{Other} & train-other-500 & 496.7 & \multirow{3}{*}{507} & 1,166 & \multirow{3}{*}{1,232}  \\
	& dev-other       & 5.3   &                      & 33    &                         \\
	& test-other      & 5.1   &                      & 33    &                         \\
	\bottomrule
\end{tabular}

	\caption{The statistics of the LibrSpeech corpus.} 
	\label{tab:libir_stat}
\end{table}

%% file: tables/8_vocab_compare.tex
\begin{table}[h]

	\centering

\begin{tabular}{c|c|c|c|c} 
	\toprule
	& \multirow{2}{*}{Official} & \multicolumn{2}{c|}{Libri-Clean} & \multicolumn{1}{l}{Libri-All}  \\ 
	\cline{3-5}
	&                           & -      & count$\ge $5                 & count$\ge $5                        \\ 
	\hline
	Vocab     &  37,622                    & 66,721 & 27,454                  & \textbf{37,622}                         \\
	Missing & -                         & 1,685  & 10,168                  & \textbf{0}                              \\
	\bottomrule
\end{tabular}

	\caption{ The vocabulary composition. }
	 
	\label{tab:vocab_compare}
\end{table}

%% file: tables/5_missing_words.tex
\begin{table}[h]

	\centering
\begin{tabular}{l|cc} 
	\toprule
	\multirow{2}{*}{Benchmark} & \multicolumn{2}{c}{Not Found Pairs}  \\
	& Paper Data & Libri-Clean                   \\ 
	\hline
MEN~\cite{test_men}          & 122   & 231   \\
MTurk-287~\cite{test_MTurk-287}    & 13    & 29    \\
MTurk-771~\cite{test_MTurk-771}    & 22    & 43    \\
Rare-Word~\cite{test_Rare-Word}    & 783   & 1027  \\
SimLex-999~\cite{test_simlex}  & 0     & 11    \\
SimVerb-3500~\cite{test_SimVerb-3500} & 126   & 118   \\
Verb-143~\cite{test_Verb-143}     & 0     & 5     \\
WS-353~\cite{test_WS_353}       & 21    & 34    \\
WS-353-REL~\cite{test_WS353REL}   & 12    & 24    \\
WS-353-SIM~\cite{test_WS353REL}   & 7     & 16    \\
YP-130~\cite{test_WS_353}       & 0     & 9     \\
	\bottomrule
\end{tabular}
	\caption{
		Not found pairs in word similarity benchmarks.  
	}
	\label{tab:missing_word}
\end{table}

%% file: tables/6_score_compare.tex
\begin{table}[h]

	\centering
\begin{tabular}{l|c|c|c|c|c} 
	\toprule
	\multirow{2}{*}{Benchmark} & \multicolumn{1}{l|}{\multirow{2}{*}{Paper Data}} & \multicolumn{2}{c|}{SGD} & \multicolumn{2}{c}{Adam}   \\ 
	\cline{3-6}
	& \multicolumn{1}{l|}{}                            & Rand Init & 500-Epoch         & Rand Init        & 500-Epoch    \\ 
	\hline
MC-30 & 0.85 & 0.18 & 0.24{\tiny ↓0.60} & 0.37 & -0.28{\tiny ↓1.13}\\
MEN & 0.62 & 0.08 & 0.08{\tiny ↓0.54} & 0.06 & 0.04{\tiny ↓0.58}\\
MTurk-287 & 0.47 & 0.00 & 0.02{\tiny ↓0.45} & 0.04 & 0.08{\tiny ↓0.39}\\
MTurk-771 & 0.52 & -0.00 & 0.03{\tiny ↓0.49} & -0.00 & 0.04{\tiny ↓0.48}\\
RG-65 & 0.79 & 0.11 & 0.18{\tiny ↓0.61} & 0.25 & 0.05{\tiny ↓0.74}\\
Rare-Word & 0.32 & 0.10 & 0.12{\tiny ↓0.21} & 0.10 & 0.04{\tiny ↓0.28}\\
SimLex-999 & 0.29 & 0.02 & -0.05{\tiny ↓0.34} & -0.06 & -0.04{\tiny ↓0.33}\\
SimVerb-3500 & 0.16 & 0.00 & -0.02{\tiny ↓0.17} & -0.03 & -0.03{\tiny ↓0.18}\\
Verb-143 & 0.32 & 0.24 & 0.29{\tiny ↓0.02} & 0.34 & 0.24{\tiny ↓0.07}\\
WS-353 & 0.51 & 0.11 & 0.14{\tiny ↓0.36} & 0.18 & 0.12{\tiny ↓0.39}\\
WS-353-REL & 0.35 & 0.10 & 0.11{\tiny ↓0.24} & 0.12 & 0.11{\tiny ↓0.24}\\
WS-353-SIM & 0.66 & 0.10 & 0.15{\tiny ↓0.51} & 0.22 & 0.16{\tiny ↓0.51}\\
YP-130 & 0.32 & 0.01 & 0.07{\tiny ↓0.25} & 0.11 & -0.01{\tiny ↓0.33}\\
	\bottomrule
\end{tabular}
	\caption{
Results on semantic benchmarks.
The ↓ represents the gap with the Paper Data.
Note that the optimizer type does not affect the `Rand Init' results.
We used different seeds for these two runs; that's why their Rand Init results are different.
} 
	\label{tab:score_compare}
\end{table}

%% file: tables/9_full_homolist.tex
	\begin{longtable}{llccc|llccc}
		\caption{The homophone list. } 
		\label{tab:homo_full_list} \\ 

		\hline 
	\multicolumn{2}{l}{Homophone Pairs} & Raw  & Official & Ours & \multicolumn{2}{l}{Homophone Pairs} & Raw & Official & Ours  \\ 
		\hline 
		\endfirsthead
		
		\hline 
	\multicolumn{2}{l}{Homophone Pairs} & Raw  & Official & Ours & \multicolumn{2}{l}{Homophone Pairs} & Raw & Official & Ours  \\ 
		\hline 
		\endhead
		
		\hline 
		\endfoot
		
		\hline 
		\endlastfoot
ad & add & 0.99 & 0.09 & 0.92 & lieu & loo & 0.98 & 0.43 & 0.91 \\
ail & ale & 0.98 & 0.49 & 0.75 & links & lynx & 1.00 & 0.24 & 0.96 \\
air & heir & 1.00 & 0.07 & 0.97 & lo & low & 0.99 & 0.22 & 0.98 \\
all & awl & - & 0.40 & - & loan & lone & 1.00 & 0.09 & 0.94 \\
allowed & aloud & 1.00 & 0.21 & 0.96 & loot & lute & 0.99 & 0.17 & 0.92 \\
alms & arms & 0.95 & 0.26 & 0.95 & made & maid & 1.00 & 0.25 & 0.98 \\
altar & alter & 0.99 & 0.13 & 0.95 & mail & male & 1.00 & 0.23 & 0.98 \\
arc & ark & 0.98 & 0.19 & 0.91 & main & mane & 0.99 & 0.17 & 0.97 \\
aren't & aunt & 0.96 & 0.44 & 0.95 & maize & maze & 0.99 & 0.25 & 0.93 \\
ate & eight & 1.00 & 0.22 & 1.00 & manna & manner & 0.97 & 0.33 & 0.89 \\
auger & augur & - & 0.51 & - & mantel & mantle & 1.00 & 0.43 & 0.95 \\
awe & oar & 0.92 & 0.26 & 0.78 & mare & mayor & 0.99 & 0.38 & 0.96 \\
axel & axle & 0.99 & 0.41 & 0.86 & marshal & martial & 1.00 & 0.60 & 0.98 \\
aye & eye & 0.99 & 0.29 & 0.89 & marten & martin & - & 0.36 & - \\
bail & bale & 0.99 & 0.36 & 0.98 & mask & masque & - & 0.51 & - \\
bait & bate & 0.98 & 0.41 & 0.95 & mean & mien & 0.98 & 0.24 & 0.90 \\
baize & bays & 0.99 & 0.50 & 0.91 & meat & meet & 1.00 & 0.13 & 0.99 \\
bald & bawled & 0.99 & 0.21 & 0.96 & medal & meddle & 0.99 & 0.11 & 0.90 \\
bard & barred & 0.98 & 0.01 & 0.96 & metal & mettle & 1.00 & 0.31 & 0.92 \\
bare & bear & 1.00 & 0.51 & 1.00 & meter & metre & 0.99 & 0.59 & 0.95 \\
bark & barque & 0.96 & 0.33 & 0.86 & might & mite & 0.99 & 0.38 & 0.96 \\
baron & barren & 1.00 & 0.07 & 0.97 & mind & mined & - & 0.21 & - \\
base & bass & 0.99 & 0.28 & 0.98 & miner & minor & 1.00 & 0.17 & 0.95 \\
bazaar & bizarre & 0.98 & 0.12 & 0.92 & missed & mist & 0.99 & 0.20 & 0.98 \\
be & bee & 0.99 & 0.35 & 0.98 & moat & mote & 0.99 & 0.46 & 0.93 \\
beach & beech & 1.00 & 0.47 & 0.97 & mode & mowed & 0.99 & 0.21 & 0.94 \\
bean & been & 0.92 & 0.29 & 0.96 & moor & more & 0.99 & 0.41 & 0.94 \\
beat & beet & - & 0.41 & - & morning & mourning & 1.00 & 0.46 & 0.99 \\
beer & bier & 0.99 & 0.24 & 0.96 & muscle & mussel & 0.99 & 0.48 & 0.95 \\
bel & bell & 0.99 & 0.29 & 0.98 & naval & navel & - & 0.34 & - \\
berry & bury & 0.99 & 0.16 & 0.97 & nay & neigh & - & 0.34 & - \\
berth & birth & 1.00 & 0.08 & 1.00 & none & nun & 0.99 & 0.35 & 0.84 \\
bitten & bittern & - & 0.63 & - & ode & owed & 0.99 & 0.30 & 0.95 \\
blew & blue & 1.00 & 0.48 & 1.00 & oh & owe & 0.97 & 0.55 & 0.82 \\
boar & bore & 0.99 & 0.39 & 0.96 & one & won & 1.00 & 0.39 & 1.00 \\
board & bored & 1.00 & 0.13 & 0.99 & pail & pale & 1.00 & 0.27 & 0.99 \\
boarder & border & 0.99 & 0.34 & 0.92 & pain & pane & 1.00 & 0.38 & 0.98 \\
bold & bowled & 0.99 & 0.50 & 0.92 & pair & pare & 0.99 & 0.35 & 0.95 \\
born & borne & 1.00 & 0.29 & 0.99 & palate & palette & 0.98 & 0.50 & 0.88 \\
bough & bow & 0.99 & 0.48 & 0.97 & pause & paws & 1.00 & 0.34 & 0.99 \\
boy & buoy & 0.96 & 0.35 & 0.88 & pea & pee & 0.93 & 0.54 & 0.87 \\
braid & brayed & - & 0.60 & - & peace & piece & 1.00 & 0.15 & 0.99 \\
brake & break & 1.00 & 0.38 & 0.99 & peak & peek & - & 0.30 & - \\
bread & bred & 1.00 & 0.24 & 0.99 & peal & peel & 0.99 & 0.21 & 0.96 \\
bridal & bridle & 1.00 & 0.36 & 0.97 & peer & pier & 0.99 & 0.33 & 0.96 \\
broach & brooch & 0.98 & 0.52 & 0.89 & plain & plane & 1.00 & 0.40 & 0.99 \\
but & butt & 0.97 & 0.37 & 0.92 & pleas & please & - & 0.42 & - \\
buy & by & 0.99 & 0.25 & 0.98 & plum & plumb & 0.99 & 0.57 & 0.96 \\
calendar & calender & 0.98 & 0.39 & 0.95 & pole & poll & 0.98 & 0.19 & 0.96 \\
canvas & canvass & 0.99 & 0.41 & 0.98 & practice & practise & 1.00 & 0.74 & 0.99 \\
cast & caste & 0.99 & 0.33 & 0.98 & praise & prays & 0.99 & 0.51 & 0.99 \\
caught & court & 0.95 & 0.11 & 0.91 & principal & principle & 1.00 & 0.48 & 1.00 \\
ceiling & sealing & 1.00 & 0.63 & 0.95 & profit & prophet & 1.00 & 0.30 & 0.99 \\
cell & sell & 0.99 & 0.06 & 0.95 & quarts & quartz & 0.99 & 0.49 & 0.95 \\
censer & censor & 0.98 & 0.35 & 0.78 & rain & reign & 1.00 & 0.18 & 0.99 \\
cent & scent & 0.99 & 0.00 & 0.98 & raise & rays & 1.00 & 0.10 & 0.99 \\
check & cheque & 0.99 & 0.61 & 0.98 & rap & wrap & 0.99 & 0.40 & 0.96 \\
chord & cord & 1.00 & 0.42 & 0.98 & raw & roar & 0.96 & 0.18 & 0.88 \\
cite & sight & 0.99 & 0.24 & 0.97 & read & reed & 0.97 & 0.28 & 0.97 \\
clew & clue & 0.99 & 0.83 & 0.96 & real & reel & 1.00 & 0.27 & 0.97 \\
climb & clime & 0.99 & 0.29 & 0.93 & reek & wreak & 0.98 & 0.33 & 0.92 \\
coarse & course & 1.00 & 0.20 & 0.99 & rest & wrest & 0.99 & 0.54 & 0.95 \\
coign & coin & - & 0.37 & - & right & rite & 1.00 & 0.19 & 0.96 \\
colonel & kernel & 0.98 & 0.02 & 0.94 & ring & wring & 0.99 & 0.48 & 0.97 \\
complacent & complaisant & - & 0.54 & - & road & rode & 1.00 & 0.66 & 0.96 \\
complement & compliment & 1.00 & 0.27 & 0.96 & roe & row & 0.95 & 0.23 & 0.80 \\
coo & coup & 0.98 & 0.35 & 0.87 & role & roll & 1.00 & 0.18 & 0.99 \\
cops & copse & 0.98 & 0.39 & 0.96 & rood & rude & - & 0.32 & - \\
council & counsel & 1.00 & 0.52 & 0.99 & root & route & 0.98 & 0.06 & 0.98 \\
creak & creek & 0.99 & 0.37 & 0.98 & rose & rows & 1.00 & 0.37 & 0.99 \\
crews & cruise & 1.00 & 0.67 & 0.95 & rote & wrote & 0.98 & 0.46 & 0.91 \\
currant & current & 0.99 & 0.11 & 0.97 & rough & ruff & 0.99 & 0.51 & 0.94 \\
dam & damn & 0.99 & 0.21 & 0.95 & rung & wrung & 1.00 & 0.33 & 0.97 \\
days & daze & 0.99 & 0.47 & 0.94 & rye & wry & 0.99 & 0.35 & 0.92 \\
dear & deer & 0.99 & 0.08 & 0.98 & sale & sail & 1.00 & 0.32 & 0.99 \\
descent & dissent & 1.00 & 0.42 & 0.98 & sane & seine & - & 0.09 & - \\
desert & dessert & 0.99 & 0.14 & 0.97 & sauce & source & 0.97 & 0.22 & 0.97 \\
dew & due & 1.00 & 0.25 & 0.98 & saw & soar & 0.96 & 0.33 & 0.92 \\
die & dye & 0.99 & 0.26 & 0.89 & scene & seen & 1.00 & 0.41 & 1.00 \\
done & dun & 0.99 & 0.21 & 0.97 & sea & see & 1.00 & 0.31 & 0.98 \\
draft & draught & 0.99 & 0.30 & 0.96 & seam & seem & 0.98 & 0.30 & 0.93 \\
dual & duel & 0.98 & 0.48 & 0.88 & sear & seer & 0.98 & 0.57 & 0.86 \\
earn & urn & 0.99 & 0.04 & 0.89 & seas & sees & 0.99 & 0.18 & 0.96 \\
ewe & yew & 0.99 & 0.45 & 0.90 & sew & so & 0.99 & 0.55 & 0.98 \\
faint & feint & 0.99 & 0.55 & 0.95 & shear & sheer & 0.99 & 0.25 & 0.94 \\
fair & fare & 1.00 & 0.42 & 0.99 & side & sighed & 1.00 & 0.31 & 0.99 \\
farther & father & 0.98 & 0.29 & 0.99 & sign & sine & - & 0.39 & - \\
feat & feet & 1.00 & 0.37 & 0.99 & slay & sleigh & 0.99 & 0.30 & 0.97 \\
few & phew & - & 0.38 & - & sole & soul & 1.00 & 0.43 & 0.98 \\
find & fined & 0.99 & 0.24 & 0.97 & some & sum & 0.99 & 0.34 & 0.99 \\
fir & fur & 1.00 & 0.32 & 0.99 & son & sun & 1.00 & 0.26 & 1.00 \\
flaw & floor & 0.97 & 0.30 & 0.87 & sort & sought & 0.96 & 0.21 & 0.96 \\
flea & flee & 0.99 & 0.36 & 0.98 & staid & stayed & 1.00 & 0.55 & 0.99 \\
floe & flow & 0.99 & 0.42 & 0.89 & stair & stare & 1.00 & 0.27 & 0.99 \\
flour & flower & 1.00 & 0.30 & 0.99 & stake & steak & 1.00 & 0.19 & 0.98 \\
for & fore & 0.95 & 0.26 & 0.99 & stalk & stork & 0.96 & 0.55 & 0.92 \\
fort & fought & 0.95 & 0.42 & 0.93 & stationary & stationery & 0.99 & 0.27 & 0.85 \\
forth & fourth & 1.00 & 0.25 & 1.00 & steal & steel & 1.00 & 0.19 & 0.99 \\
foul & fowl & 1.00 & 0.36 & 0.97 & stile & style & 0.99 & 0.21 & 0.89 \\
franc & frank & 0.98 & 0.31 & 0.95 & storey & story & 0.99 & 0.37 & 0.95 \\
freeze & frieze & - & 0.38 & - & straight & strait & 1.00 & 0.39 & 0.97 \\
furs & furze & - & 0.56 & - & sweet & suite & 1.00 & 0.03 & 0.99 \\
gait & gate & 1.00 & 0.31 & 0.99 & tacks & tax & 0.98 & 0.25 & 0.96 \\
gamble & gambol & - & 0.41 & - & tale & tail & 1.00 & 0.23 & 1.00 \\
gild & guild & - & 0.41 & - & taught & taut & 0.99 & 0.12 & 0.93 \\
gilt & guilt & 1.00 & 0.21 & 0.99 & team & teem & - & 0.28 & - \\
gnaw & nor & 0.95 & 0.38 & 0.84 & tear & tier & 0.98 & 0.24 & 0.92 \\
grate & great & 1.00 & 0.15 & 0.97 & teas & tease & 0.99 & 0.35 & 0.97 \\
greys & graze & 0.98 & 0.65 & 0.90 & tern & turn & - & 0.32 & - \\
grisly & grizzly & 0.99 & 0.64 & 0.88 & there & their & 0.99 & 0.34 & 0.98 \\
groan & grown & 0.99 & 0.25 & 0.95 & threw & through & 0.99 & 0.39 & 0.99 \\
guessed & guest & 1.00 & 0.33 & 0.99 & throes & throws & 0.99 & 0.31 & 0.95 \\
hail & hale & 0.99 & 0.34 & 0.97 & throne & thrown & 1.00 & 0.30 & 0.96 \\
hair & hare & 1.00 & 0.32 & 0.98 & thyme & time & 0.99 & 0.25 & 0.97 \\
hall & haul & 1.00 & 0.13 & 0.93 & tide & tied & 1.00 & 0.10 & 0.99 \\
hart & heart & 0.99 & 0.23 & 0.94 & tire & tyre & 0.99 & 0.21 & 0.90 \\
haw & hoar & 0.91 & 0.12 & 0.71 & to & too & 0.97 & 0.63 & 0.97 \\
hay & hey & 0.98 & 0.37 & 0.92 & toad & toed & 0.98 & 0.55 & 0.96 \\
he'd & heed & 0.99 & 0.39 & 0.98 & told & tolled & 0.98 & 0.33 & 0.96 \\
heal & heel & 1.00 & 0.37 & 0.95 & ton & tun & - & 0.55 & - \\
hear & here & 1.00 & 0.58 & 0.98 & vain & vane & 0.99 & 0.19 & 0.96 \\
heard & herd & 0.99 & 0.32 & 0.97 & vale & veil & 1.00 & 0.65 & 0.98 \\
hew & hue & 0.99 & 0.34 & 0.94 & vial & vile & 0.99 & 0.33 & 0.94 \\
hi & high & 0.98 & 0.18 & 0.98 & wain & wane & 0.98 & 0.73 & 0.86 \\
higher & hire & 1.00 & 0.27 & 0.98 & waist & waste & 1.00 & 0.13 & 0.99 \\
him & hymn & 0.99 & 0.28 & 0.98 & wait & weight & 1.00 & 0.21 & 0.99 \\
ho & hoe & 0.98 & 0.40 & 0.96 & waive & wave & 0.99 & 0.22 & 0.95 \\
hoard & horde & 0.99 & 0.47 & 0.89 & war & wore & 0.99 & 0.23 & 0.96 \\
hoarse & horse & 1.00 & 0.21 & 0.99 & ware & wear & 0.99 & 0.32 & 0.97 \\
hour & our & 0.98 & 0.28 & 0.94 & warn & worn & 1.00 & 0.02 & 0.98 \\
idle & idol & 1.00 & 0.46 & 0.97 & watt & what & - & 0.52 & - \\
in & inn & 0.98 & 0.38 & 0.95 & wax & whacks & - & 0.48 & - \\
it's & its & 0.99 & 0.20 & 0.95 & way & weigh & 0.99 & 0.36 & 0.96 \\
key & quay & 0.97 & 0.33 & 0.95 & we & wee & 0.97 & 0.21 & 0.96 \\
knave & nave & 0.99 & 0.06 & 0.90 & we'd & weed & 0.96 & 0.26 & 0.94 \\
knead & need & 0.99 & 0.51 & 0.98 & weak & week & 1.00 & 0.10 & 1.00 \\
knew & new & 1.00 & 0.31 & 0.99 & weal & we'll & 0.94 & 0.15 & 0.90 \\
knight & night & 1.00 & 0.26 & 0.99 & wean & ween & - & 0.57 & - \\
knob & nob & - & 0.59 & - & weather & whether & 0.99 & 0.21 & 0.95 \\
knot & not & 1.00 & 0.26 & 0.98 & weir & we're & - & 0.08 & - \\
know & no & 1.00 & 0.63 & 0.97 & were & whirr & 0.96 & 0.27 & 0.83 \\
knows & nose & 1.00 & 0.18 & 0.99 & which & witch & 0.99 & 0.28 & 0.94 \\
lac & lack & - & 0.10 & - & whig & wig & 0.99 & 0.16 & 0.96 \\
lain & lane & 0.99 & 0.30 & 0.96 & whine & wine & 1.00 & 0.13 & 0.97 \\
laps & lapse & - & 0.32 & - & whirl & whorl & 0.97 & 0.37 & 0.78 \\
larva & lava & - & 0.45 & - & whirled & world & 1.00 & 0.26 & 0.96 \\
law & lore & 0.96 & 0.33 & 0.89 & whit & wit & 0.99 & 0.50 & 0.96 \\
lea & lee & 0.98 & 0.53 & 0.89 & white & wight & 0.99 & 0.32 & 0.95 \\
lead & led & 0.94 & 0.59 & 0.97 & who's & whose & 0.99 & 0.26 & 0.99 \\
lessen & lesson & 1.00 & 0.26 & 0.92 & woe & whoa & 0.97 & 0.23 & 0.90 \\
levee & levy & 0.98 & 0.45 & 0.82 & wood & would & 0.98 & 0.26 & 0.96 \\
liar & lyre & 0.99 & 0.31 & 0.93 & yoke & yolk & 0.99 & 0.32 & 0.88 \\
licence & license & 0.99 & 0.72 & 0.94 & you'll & yule & 0.95 & 0.48 & 0.91 \\
lie & lye & 0.97 & 0.47 & 0.71 \\

	\end{longtable}

%% file: ReviewTemplate.bbl
\begin{thebibliography}{10}\itemsep=-1pt

\bibitem{test_WS353REL}
Eneko Agirre, Enrique Alfonseca, Keith Hall, Jana Kravalova, Marius Pasca, and
  Aitor Soroa.
\newblock A study on similarity and relatedness using distributional and
  wordnet-based approaches.
\newblock 2009.

\bibitem{test_Verb-143}
Simon Baker, Roi Reichart, and Anna Korhonen.
\newblock An unsupervised model for instance level subcategorization
  acquisition.
\newblock In {\em EMNLP}, pages 278--289, 2014.

\bibitem{lm2000}
Yoshua Bengio, R{\'e}jean Ducharme, and Pascal Vincent.
\newblock A neural probabilistic language model.
\newblock {\em Advances in neural information processing systems}, 13, 2000.

\bibitem{test_men}
Elia Bruni, Gemma Boleda, Marco Baroni, and Nam-Khanh Tran.
\newblock Distributional semantics in technicolor.
\newblock In {\em Proceedings of the 50th Annual Meeting of the Association for
  Computational Linguistics (Volume 1: Long Papers)}, pages 136--145, 2012.

\bibitem{A43_chen2018phonetic}
Yi-Chen Chen, Sung-Feng Huang, Chia-Hao Shen, Hung-Yi Lee, and Lin-Shan Lee.
\newblock Phonetic-and-semantic embedding of spoken words with applications in
  spoken content retrieval.
\newblock In {\em 2018 IEEE Spoken Language Technology Workshop (SLT)}, pages
  941--948. IEEE, 2018.

\bibitem{s2v_early2017}
Yu-An Chung and James Glass.
\newblock Learning word embeddings from speech.
\newblock {\em arXiv preprint arXiv:1711.01515}, 2017.

\bibitem{speech2vec2018}
Yu-An Chung and James Glass.
\newblock Speech2vec: A sequence-to-sequence framework for learning word
  embeddings from speech.
\newblock In {\em INTERSPEECH}, 2018.

\bibitem{A27_alignment_speech_text}
Yu-An Chung, Wei-Hung Weng, Schrasing Tong, and James Glass.
\newblock Unsupervised cross-modal alignment of speech and text embedding
  spaces.
\newblock {\em Advances in neural information processing systems}, 31, 2018.

\bibitem{A7_chung2019towards}
Yu-An Chung, Wei-Hung Weng, Schrasing Tong, and James Glass.
\newblock Towards unsupervised speech-to-text translation.
\newblock In {\em ICASSP 2019-2019 IEEE International Conference on Acoustics,
  Speech and Signal Processing (ICASSP)}, pages 7170--7174. IEEE, 2019.

\bibitem{benchmark13}
Manaal Faruqui and Chris Dyer.
\newblock Community evaluation and exchange of word vectors at wordvectors.
  org.
\newblock In {\em Proceedings of 52nd Annual Meeting of the Association for
  Computational Linguistics: System Demonstrations}, pages 19--24, 2014.

\bibitem{test_SimVerb-3500}
Daniela Gerz, Ivan Vuli{\'c}, Felix Hill, Roi Reichart, and Anna Korhonen.
\newblock Simverb-3500: A large-scale evaluation set of verb similarity.
\newblock {\em arXiv preprint arXiv:1608.00869}, 2016.

\bibitem{test_MTurk-771}
Guy Halawi, Gideon Dror, Evgeniy Gabrilovich, and Yehuda Koren.
\newblock Large-scale learning of word relatedness with constraints.
\newblock In {\em Proceedings of the 18th ACM SIGKDD international conference
  on Knowledge discovery and data mining}, pages 1406--1414, 2012.

\bibitem{test_simlex}
Felix Hill, Roi Reichart, and Anna Korhonen.
\newblock Simlex-999: Evaluating semantic models with (genuine) similarity
  estimation.
\newblock {\em Computational Linguistics}, 41(4):665--695, 2015.

\bibitem{luong2015attention}
Minh-Thang Luong, Hieu Pham, and Christopher~D Manning.
\newblock Effective approaches to attention-based neural machine translation.
\newblock {\em arXiv preprint arXiv:1508.04025}, 2015.

\bibitem{test_Rare-Word}
Minh-Thang Luong, Richard Socher, and Christopher~D Manning.
\newblock Better word representations with recursive neural networks for
  morphology.
\newblock In {\em Proceedings of the seventeenth conference on computational
  natural language learning}, pages 104--113, 2013.

\bibitem{montreal}
Michael McAuliffe, Michaela Socolof, Sarah Mihuc, Michael Wagner, and Morgan
  Sonderegger.
\newblock Montreal forced aligner: Trainable text-speech alignment using kaldi.
\newblock In {\em Interspeech}, volume 2017, pages 498--502, 2017.

\bibitem{word2vec}
Tomas Mikolov, Ilya Sutskever, Kai Chen, Greg~S Corrado, and Jeff Dean.
\newblock Distributed representations of words and phrases and their
  compositionality.
\newblock {\em Advances in neural information processing systems}, 26, 2013.

\bibitem{A18_CAWE}
Shruti Palaskar, Vikas Raunak, and Florian Metze.
\newblock Learned in speech recognition: Contextual acoustic word embeddings.
\newblock In {\em ICASSP 2019-2019 IEEE International Conference on Acoustics,
  Speech and Signal Processing (ICASSP)}, pages 6530--6534. IEEE, 2019.

\bibitem{librispeech}
Vassil Panayotov, Guoguo Chen, Daniel Povey, and Sanjeev Khudanpur.
\newblock Librispeech: an asr corpus based on public domain audio books.
\newblock In {\em 2015 IEEE international conference on acoustics, speech and
  signal processing (ICASSP)}, pages 5206--5210. IEEE, 2015.

\bibitem{test_MTurk-287}
Kira Radinsky, Eugene Agichtein, Evgeniy Gabrilovich, and Shaul Markovitch.
\newblock A word at a time: computing word relatedness using temporal semantic
  analysis.
\newblock In {\em Proceedings of the 20th international conference on World
  wide web}, pages 337--346, 2011.

\bibitem{wav2vec2019}
Steffen Schneider, Alexei Baevski, Ronan Collobert, and Michael Auli.
\newblock wav2vec: Unsupervised pre-training for speech recognition.
\newblock {\em arXiv preprint arXiv:1904.05862}, 2019.

\bibitem{A76_Audio2Vec}
Marco Tagliasacchi, Beat Gfeller, F{\'e}lix de~Chaumont Quitry, and Dominik
  Roblek.
\newblock Self-supervised audio representation learning for mobile devices.
\newblock {\em arXiv preprint arXiv:1905.11796}, 2019.

\bibitem{question_answer}
Stefanie Tellex, Boris Katz, Jimmy Lin, Aaron Fernandes, and Gregory Marton.
\newblock Quantitative evaluation of passage retrieval algorithms for question
  answering.
\newblock In {\em Proceedings of the 26th annual international ACM SIGIR
  conference on Research and development in informaion retrieval}, pages
  41--47, 2003.

\bibitem{ner}
Joseph Turian, Lev Ratinov, and Yoshua Bengio.
\newblock Word representations: a simple and general method for semi-supervised
  learning.
\newblock In {\em Proceedings of the 48th annual meeting of the association for
  computational linguistics}, pages 384--394, 2010.

\bibitem{MDS2003}
Florian Wickelmaier.
\newblock An introduction to mds.
\newblock {\em Sound Quality Research Unit, Aalborg University, Denmark},
  46(5):1--26, 2003.

\bibitem{test_WS_353}
Dongqiang Yang and David~MW Powers.
\newblock {\em Verb similarity on the taxonomy of WordNet}.
\newblock Masaryk University, 2006.

\bibitem{A8_punctuation_prediction}
Jiangyan Yi and Jianhua Tao.
\newblock Self-attention based model for punctuation prediction using word and
  speech embeddings.
\newblock In {\em ICASSP 2019-2019 IEEE International Conference on Acoustics,
  Speech and Signal Processing (ICASSP)}, pages 7270--7274. IEEE, 2019.

\end{thebibliography}
